\documentclass{article}
\usepackage{amsmath,graphicx,hyperref}

\usepackage{amssymb}
\usepackage{multirow}
\usepackage{multirow}
\usepackage{booktabs}
\usepackage{makecell}

\usepackage[preprint]{spconf} 
\copyrightnotice{\copyright\ IEEE 2026}
\toappear{To appear in {\it Proc.\ ICASSP 2026,
   May 04-08, 2026, Barcelona, Spain}}

\title{A DyL-Unet framework based on dynamic learning for Temporally Consistent Echocardiographic Segmentation}
\twoauthors
 {Jierui Qu}
	{National University of Singapore\\
	College of Design and Engineering\\
	Singapore 117575, Singapore}
 {Jianchun Zhao\sthanks{Corresponding author, email: zhao\_jianchun@stu.xjtu.edu.cn}}
	{Xi'an Jiaotong University\\
	School of Electrical Engineering\\
	Xi'an 710049, China.}
\begin{document}
\ninept
\maketitle
\begin{abstract}
Accurate segmentation of cardiac anatomy in echocardiography is essential for cardiovascular diagnosis and treatment. Yet echocardiography is prone to deformation and speckle noise, causing frame-to-frame segmentation jitter. Even with high accuracy in single-frame segmentation, temporal instability can weaken functional estimates and impair clinical interpretability. To address these issues, we propose DyL-UNet, a dynamic learning-based temporal consistency U-Net segmentation architecture designed to achieve temporally stable and precise echocardiographic segmentation. The framework constructs an Echo-Dynamics Graph (EDG) through dynamic learning to extract dynamic information from videos. DyL-UNet incorporates multiple Swin-Transformer-based encoder-decoder branches for processing single-frame images. It further introduces Cardiac Phase-Dynamics Attention (CPDA) at the skip connections, which uses EDG-encoded dynamic features and cardiac-phase cues to enforce temporal consistency during segmentation. Extensive experiments on the CAMUS and EchoNet-Dynamic datasets demonstrate that DyL-UNet maintains segmentation accuracy comparable to existing methods while achieving superior temporal consistency, providing a reliable solution for automated clinical echocardiography.
\end{abstract}
\begin{keywords}
Dynamic Learning, Echocardiography, cardiac segmentation, video segmentation
\end{keywords}
\section{Introduction}
\label{sec:intro}
Cardiovascular disease is the leading cause of death worldwide~\cite{chen2020deep}. Echocardiography serves as an indispensable clinical tool for assessing cardiovascular function, enabling the evaluation and quantification of cardiac chamber size and function. Currently, clinical assessments largely rely on echocardiographers visually identifying cardiac cycle phases and manually delineating chamber borders~\cite{liu2019deep}-a process that is time-consuming, subjective, and prone to errors~\cite{koh2017comprehensive}. Consequently, there is an urgent need for a highly accurate automated segmentation method for echocardiographic images.

In recent years, the rapid advancement of deep learning has significantly propelled research in echocardiographic segmentation. Early studies primarily focused on single frames, concentrating on segmentation during static end-diastolic/end-systolic (ED/ES) frames~\cite{zhang2018fully,moradi2019mfpunet}. Subsequently, some researchers extended this image-level segmentation approach to video data, performing independent inference for each frame~\cite{ouyang2020videobased,duffy2022highthroughput}. While these approaches enabled fully automated assessment of single-frame cardiac structures, they neglected temporal dependencies between consecutive frames. To address this limitation, subsequent studies introduced temporal modeling: one class of methods incorporated temporal units into 2D/3D networks~\cite{li2019recurrent,petit2025echodfkd}, while another employed 2D+t/3D+t convolutions to jointly extract spatio-temporal features within fixed time windows~\cite{hasan2025motionenhanced}. However, these approaches often rely on fixed-length windows or local temporal memory, making it challenging to fully characterize cardiac dynamics under complex conditions such as arrhythmia, leading to segmentation bias.

Dynamic learning is a novel machine learning algorithm proposed by Wang~\cite{wang2007deterministic,wang2018deterministic} for dynamic analysis and modeling of nonlinear systems~\cite{wu2019deterministic,yuan2014design}. Research indicates that this algorithm can be applied to the dynamic analysis of complex nonlinear cardiac systems for extracting dynamical features from electrocardiograms~\cite{wang2016new}. ECG signals exhibit distinct nonlinear dynamic characteristics. The dynamic learning algorithm identifies abnormal patterns in cardiac electrical activity by constructing feature descriptors, incrementally modeling descriptor sequences using RBF networks, and analyzing prediction residuals. This process generates dynamic maps characterizing the spatiotemporal evolution of cardiac electrical activity~\cite{deng2017cardiodynamicsgram}. Although echocardiography and ECG differ in acquisition methods, both involve temporal analysis of cardiac dynamic processes: ECG reflects temporal changes in the cardiac conduction system, while echocardiography reveals the temporal motion patterns of myocardial contraction and relaxation. This similarity in temporal dynamic characteristics provides a theoretical foundation for applying dynamic learning methods to echocardiogram analysis.

Inspired by the successful application of dynamic learning in electrocardiogram analysis, this study proposes extending the core principles of dynamic learning to echocardiographic segmentation tasks by extracting the Echo-dynamics graph (EDG) as a dynamic feature of cardiac motion in echocardiograms. Our designed DyL-UNet architecture integrates video sequences, EDG dynamic features, and cardiac phase information to achieve temporally consistent and precise segmentation of cardiac chambers. The main contributions of this study are as follows:

\begin{figure*}[htb]
\centering
\centerline{\includegraphics[width=15cm]{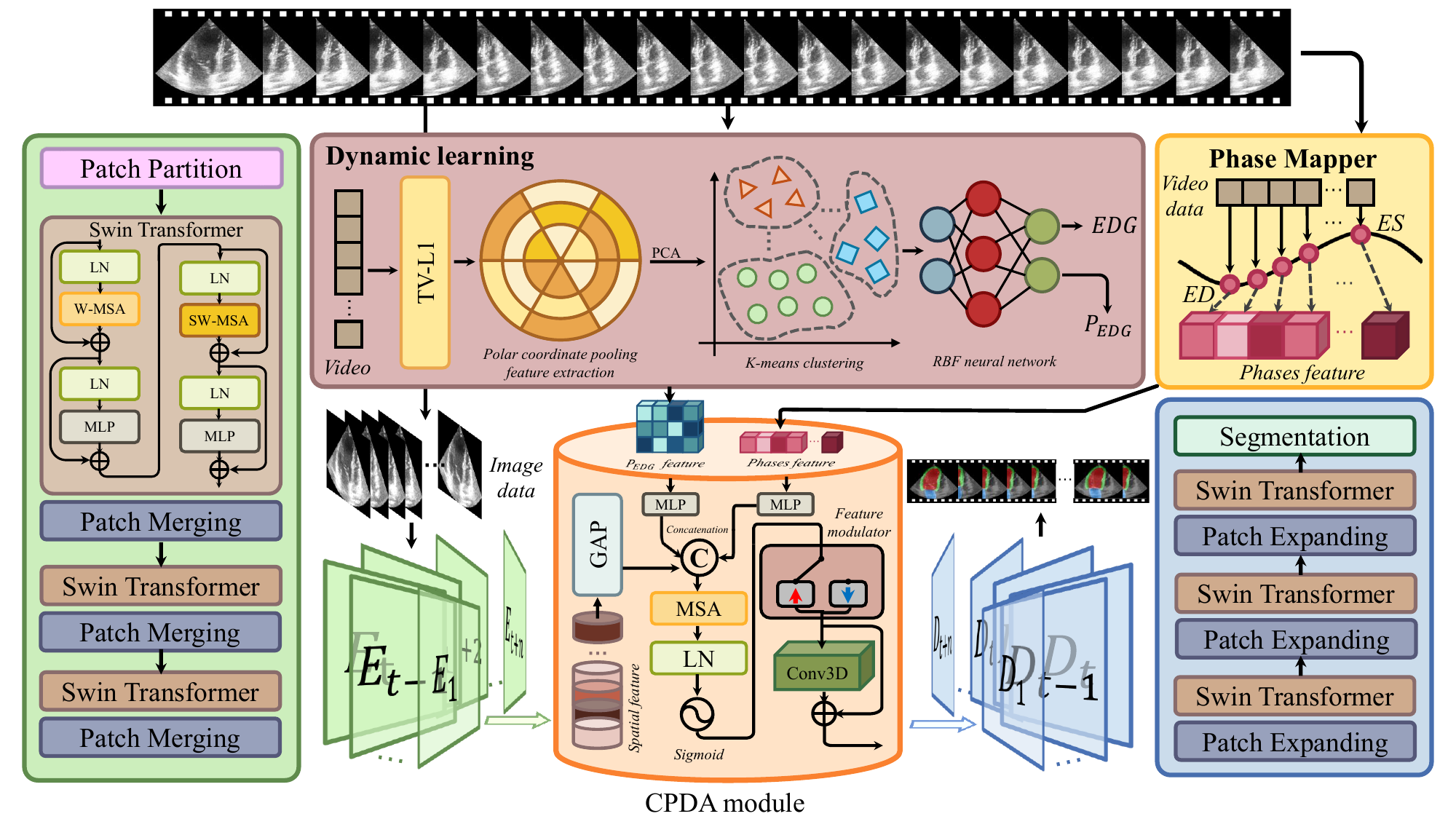}}
\caption{The overall of the DyL-Unet framework, which consists of a dynamic learning part, a CPDA module and a Multi-branch sequence encoder-decoder network based on swin transformer.}
\label{fig:overall}
\end{figure*}

\begin{itemize}
\item[$\bullet$] Extends dynamic learning to the field of echocardiography, capturing nonlinear dynamic features of cardiac motion within echocardiographic sequences to provide dynamic features for temporally coherent segmentation.
\item[$\bullet$] Designed the DyL-Unet framework with a Cardiac Phase-Dynamics Attention (CPDA) module, integrating echocardiographic spatial information, dynamic features, and phase information for temporally consistent segmentation of cardiac structures.
\item[$\bullet$] Experiments demonstrate that the proposed method achieves comparable segmentation accuracy to existing approaches while exhibiting superior temporal consistency.
\end{itemize}

\section{Method}
\label{sec:format}
\subsection{Overall}

The overall architecture of the proposed DyL-Unet based on dynamic learning is shown in Figure~\ref{fig:overall}. The framework consists of dynamic learning, a CPDA module, and a multi-branch sequence encoding-decoding network based on Swin Transformer~\cite{cao2023swinunet}. For the input echocardiogram sequence $\{I_{t}\}_{t=1}^{T}\in \mathbb{R}^{T\times H\times W\times C}$, dynamic learning is first applied to extract dynamic features of the cardiac cycle, yielding the EDG, and generating low-dimensional dynamic features $P_{EDG}$. The encoding stage employs a frame-by-frame processing strategy, where each frame $I_t$ in the sequence is independently processed through the encoder to extract multi-scale spatial feature representations $\{X_{Spatial}^{(t,l)}\}_{l=1}^{4}$. The Swin Transformer encoder follows a hierarchical window attention mechanism, performing downsampling at each level via Patch Merging layers. The decoder path mirrors the encoder symmetrically, employing Patch Expanding layers for upsampling. In the skip connections between encoder and decoder, we introduce the CPDA module. This module fuses the temporally stacked skip features $X_{Spatial}^{(l)}$ with cardiac phase information $\varphi$ and $P_{EDG}$. Through a multi-head self-attention mechanism, it achieves time-aware feature enhancement, ensuring segmentation accuracy and temporal consistency.

\begin{table*}[htbp]

\caption{Quantitative comparison of the proposed method with sota methods on the Camus and Echonet-dynamic datasets.}\label{table:comparison}
\centering
\renewcommand{\arraystretch}{1.15}
\begin{tabular}{ccccccccc}
\Xhline{1.2pt}
\multirow{3}{*}{Method} & \multicolumn{6}{c}{CAMUS} & \multicolumn{2}{c}{EchoNet-Dynamic} \\ 
\cmidrule(lr){2-7}\cmidrule(lr){8-9}
                        & \multicolumn{4}{c}{Dice($\uparrow$)} & \multirow{2}{*}{HD95($\downarrow$)} & \multirow{2}{*}{Average TCD($\downarrow$)} & \multirow{2}{*}{Dice($\uparrow$)} & \multirow{2}{*}{HD95($\downarrow$)}\\
                        \cmidrule(lr){2-5}
                        & LV & LVM & LA & Average &  &  &  & \\
\hline
PKEcho-Net & 93.98 & 87.89 & 91.75 & 91.21 & 3.91 & 0.0090 & 92.46 & 3.64\\
DSA & 94.61 & 88.29 & 92.57 & 91.82 & 4.02 & 0.0074 & 93.40 & 3.32\\
BeU2-Net & 93.84 & 87.12 & 92.26 & 91.07 & 4.67 & 0.0091 & 92.42 & 3.70\\
Echo-ODE & 93.79 & 87.60 & 90.42 & 90.60 & 5.49 & 0.0069 & 92.71 & 4.67\\
NCM-Net & 94.66 & 88.41 & 92.77 & 91.94 & 3.27 & 0.0069 & 93.18 & 2.66\\
Ours & 94.93 & 88.55 & 93.04 & 92.17 & 3.90 & 0.0062 & 92.94 & 3.59\\
\Xhline{1.2pt}
\end{tabular}
\end{table*}

\subsection{Dynamic learning}
Dynamic learning is a machine learning approach suitable for dynamic environments, capable of effectively modeling nonlinear dynamic systems to reveal their intrinsic dynamical patterns. The heart, as a naturally complex nonlinear dynamic system, generates echocardiographic sequences that can be regarded as the external manifestation of the system's periodic non-stationary signals. These echocardiographic sequences exist as discrete time series. By applying dynamic learning algorithms to model their dynamics, we extract the inter-frame dynamic features within the sequences. First, motion descriptors are extracted based on the optical flow field between consecutive frames. For the echocardiographic sequence $\{I_t\}_{t=1}^T$, the optical flow field $Flow_{t\rightarrow t+1}$ between adjacent frames is computed. A polar coordinate pooling strategy is employed to capture the radial and tangential motion patterns of the heart. Centering on the image center as the pole, the image is divided into $R\times TH$ annular sectors. For each sector $(r,\theta)$, the raw motion descriptor $d_{t,r,\theta}$ is extracted, containing the radial and tangential components of optical flow, grayscale information, and their statistical features. Principal component analysis is then applied to reduce the dimensionality of the raw descriptor to a low-dimensional representation:
\begin{equation}
z_t=\mathrm{PCA}(\text{StandardScale}(d_t))
\end{equation}
Based on dynamic learning algorithms, radial basis function networks are employed to model the dynamic characteristics of low-dimensional descriptor sequences. The cardiac cycle can be regarded as generated by the following dynamic system:
\begin{equation}
\dot{z}=F(z(t))
\end{equation}
where $F(z(t))$ denotes the unknown nonlinear dynamic function. Using sampled frame data, an RBF neural network is employed to achieve a locally precise approximation of the nonlinear dynamics in the cardiac sequence:
\begin{equation}
\Delta z_t=z_{t+1}-z_t=\sum_{i=1}^Mw_i\phi(||z_t-c_i||^2)
\end{equation}
The RBF centers $\{\mathbf{c}_i\}_{i=1}^{M}$ are obtained through K-means clustering, where $\phi(\cdot)$ denotes the Gaussian kernel function. Based on dynamic learning theory, the RBF neural network and weight update algorithm achieve a locally precise approximation of the nonlinear dynamics of the cardiac sequence. The RBF neuron subvectors satisfy partial continuity conditions within small neighborhoods along the cyclic sampling trajectory, ensuring exponential convergence of the radial basis function network weights. Finally, a dynamic energy representation is generated by computing the prediction residual-weighted basis function responses:
\begin{equation}
E_t=\Phi(z_t)\odot||\hat{\Delta z}_t-\Delta z_t||_2
\end{equation}
where $\Phi(z_t)$ represents the response vector of the RBF basis function. $E_t$ is remapped onto a sector-shaped region to obtain EDG. Secondary dimensionality reduction is applied to $E_t$, yielding the final low-dimensional dynamic feature $P_{EDG}$. This feature effectively encodes the intrinsic dynamic patterns of the cardiac cycle, providing crucial dynamic characteristics for subsequent segmentation.

\subsection{Cardiac Phase-Dynamics Attention}
The CPDA module enhances temporal consistency in segmentation results by integrating spatial features, cardiac phase information, and dynamic feature, while leveraging a multi-head self-attention mechanism to capture temporal dependencies between frames.

For the input spatial feature $X_{Spatial}^{(l)}$, adaptive average pooling is first applied to obtain $F_{pool}$. Subsequently, the cardiac phase information $\varphi$ and dynamic feature $P_{EDG}$, derived from ED/ES frames via linear estimation, are respectively encoded through multi-layer perceptrons to yield $F_{phase}$ and $F_{EDG}$. Multimodal information is fused into a unified temporal token $F_{fused}$ via feature concatenation.
This fused token is then input into a multi-head self-attention module for modeling temporal dependencies:

\begin{equation}
    F_{attn}=\text{MultiHeadAttention}(F_{fused},F_{fused},F_{fused})
\end{equation}

Attention outputs are mapped back to the feature space via linear projection and processed through a Sigmoid activation function to generate the channel modulation factor $S$. Spatial features are enhanced through residual modulation and 3D convolutions to strengthen the representation of spatio-temporal features:

\begin{equation}
    X_{mod}=X_{Spatial}^{(l)}\odot(1+\alpha\cdot(2S-1))
\end{equation}

\begin{equation}
    X_{enhanced}^{(l)}=0.5\cdot X_{mod}+0.5\cdot\mathrm{Conv}3\mathrm{D}(X_{mod})
\end{equation}

Through this design, the CPDA module effectively utilizes cardiac phase constraints and dynamic features to guide temporal modeling while preserving spatial details, achieving time-aware feature enhancement. This provides critical support for accurate and temporally consistent cardiac structure segmentation.

\section{Experience}
\subsection{Datasets}
This study utilized the publicly available CAMUS dataset and the EchoNet-dynamic dataset. The CAMUS dataset~\cite{leclerc2019deep} is a fully annotated dataset comprising 2D echocardiograms from 500 patients. This study employed its apical four-chamber (4CH) view video sequences, which include segmentation masks for the left ventricle (LV), left ventricular myocardium (LVM), and left atrium (LA). The EchoNet-Dynamic~\cite{ouyang2019echonetdynamic} dataset comprises 10,030 apical four-chamber view videos. Each video provides the area of the left ventricle as an integral, with annotations provided only for two frames:ED and ES. For EchoNet-Dynamic, we evaluate performance exclusively on the ED/ES frames.

\subsection{Comparison with state-of-the-art methods}

To demonstrate the effectiveness of the DyL-UNet architecture, we selected eight state-of-the-art (SOTA) methods for comparison, including advanced approaches such as PKEcho-Net~\cite{wu2023superefficient}, DSA~\cite{lin2024dynamicguideda}, BeU$^2$-Net~\cite{meng2025boundaryenhanced}, Echo-ODE~\cite{lu2025echoodea}, and NCM-Net~\cite{deng2025echocardiographya}, as shown in Table~\ref{table:comparison}. Evaluation metrics encompass the Dice Similarity Coefficient (Dice), Hausdorff Distance-95\% (HD95), and Temporal Consistency of Dice (TCD)~\cite{lu2025echoodea}. As shown in Table~\ref{table:comparison}, our method achieves an average Dice coefficient of 92.17\% on the CAMUS dataset, surpassing the second-best method NCM-Net (91.94\%) by 0.23 percentage points. More importantly, it demonstrates exceptional temporal consistency with an average TCD of 0.0062, reducing the error by approximately 10.1\% compared to SOTA methods. This fully validates the effectiveness of the dynamic features $P_{EDG}$ and CPDA module in maintaining segmentation temporal stability. On the EchoNet-Dynamic dataset, our method also demonstrates strong generalization capabilities, achieving a Dice coefficient of 92.94\% and an HD95 of 3.59, maintaining comparable segmentation accuracy levels with other methods. While it may slightly underperform on certain individual metrics compared to specific approaches, our method achieves a better balance between segmentation accuracy and temporal consistency.
The results confirmed the superiority of our method, with the output segmentation masks exhibiting smoother and more consistent temporal continuity. This reduction in interframe segmentation jitter provides a reliable foundation for accurate calculation of cardiac function parameters.
\subsection{Visualization of results}
\begin{figure}[!t]
\centering
\centerline{\includegraphics[width=8.5cm]{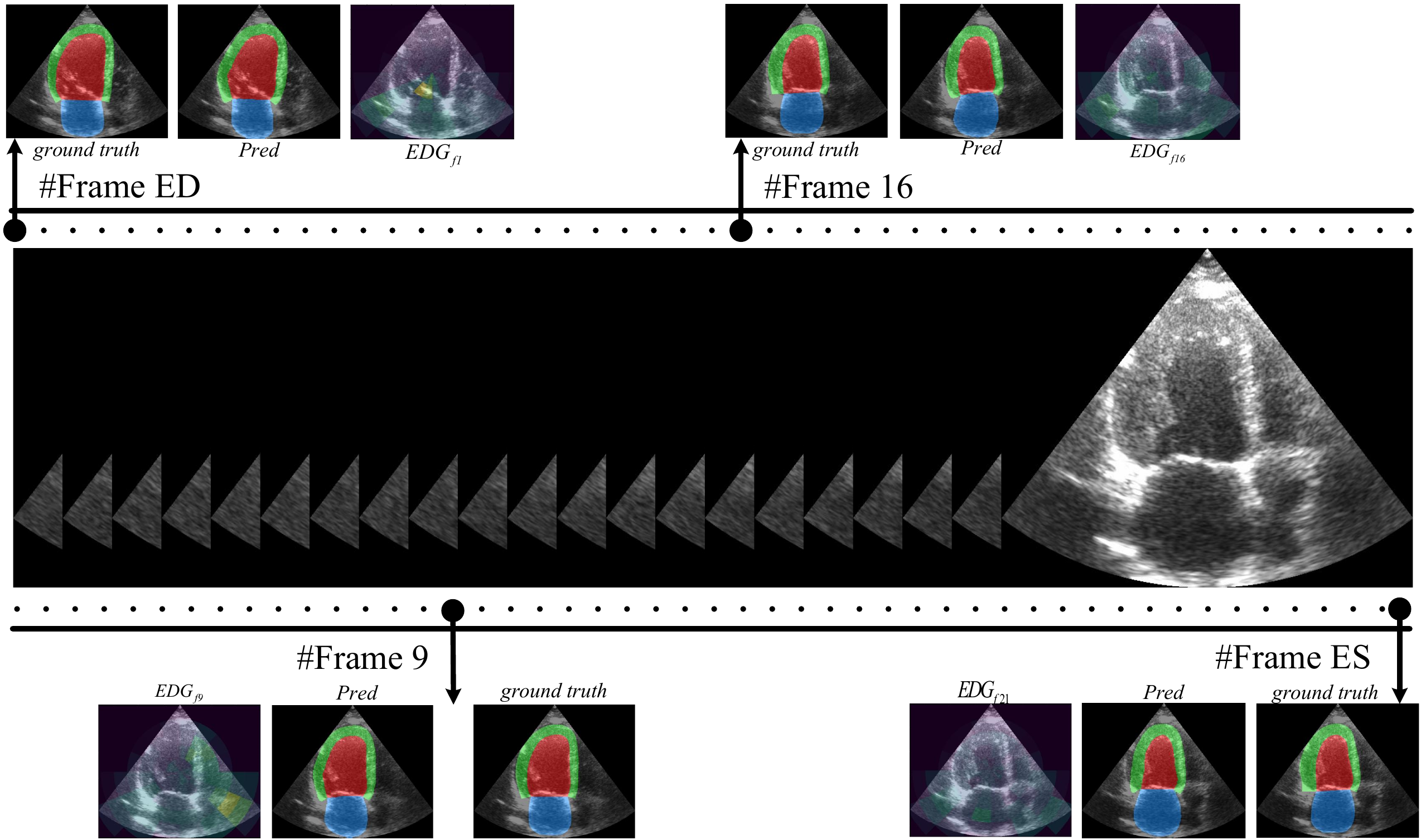}}
\caption{Ground truth, segmentation results and EDG visualization Images of an echocardiogram video. The red, green, and blue masks refer to LV, LVM, and LA cardiac structures. In EDG images, darker colors (more yellowish) indicate more pronounced dynamic features in that region.}
\label{fig:visual}
\end{figure}

To more intuitively demonstrate the segmentation performance of DyL-UNet and the mechanism of its dynamic features, we visualized the results as shown in Figure~\ref{fig:visual}. The visualizations reveal that our method accurately identifies cardiac chamber structures throughout the entire cardiac cycle, with predicted masks exhibiting high concordance with ground truth annotations. The EDG sector distribution map presented concurrently in Figure~\ref{fig:visual} reveals the spatial distribution pattern of the dynamic features. It can be observed that during the ED frame, as systole commences, localized wall segments exhibit prominent radial inward motion first, manifesting as focal hotspots. By mid-systole, contraction propagates from the endocardial layer toward the middle and outer layers, forming continuous high-energy arc bands across multiple adjacent sectors. At the end-systole (ES) frame, as the left ventricular cavity reaches its minimum volume, the dynamic characteristics diminish. The effective capture of this spatiotemporal dynamic pattern provides crucial prior information for the CPDA module, enabling it to better understand the physiological movement patterns of the heart and thereby achieve more accurate and temporally consistent segmentation results.

\subsection{Ablation Studies}
\begin{figure}[!t]
\centering
\centerline{\includegraphics[width=8.5cm]{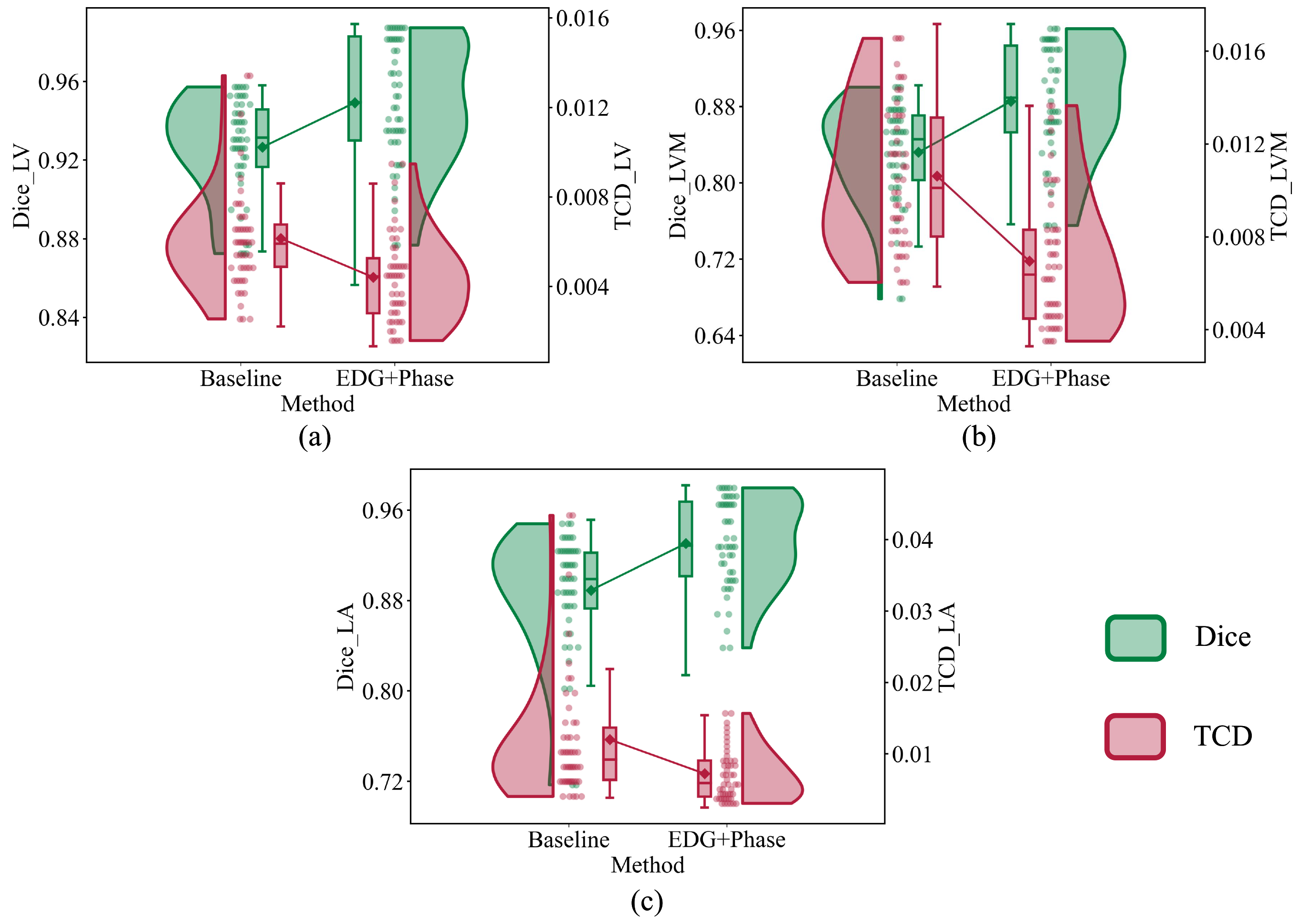}}
\caption{Comparison of segmentation results from ablation experiments. Figures (a), (b), and (c) show the Dice scores and TCD values for segmenting various cardiac chambers using the baseline model and the complete DyL-Unet framework.}
\label{fig:ablation}
\end{figure}

To validate the effectiveness of each component in DyL-UNet, we conducted ablation experiments to reveal the contributions of the dynamic learning and CPDA modules. We designed three distinct configurations for comparison: the baseline method (without phase information and dynamic features), DyL-UNet using only dynamic features (w/o phase), and the complete DyL-UNet framework. Experimental results are shown in Table~\ref{table:ablation} and Figure~\ref{fig:ablation}. As shown in Table~\ref{table:ablation}, the baseline method achieves only 88.26\% Dice coefficient and 0.0096 TCD value, demonstrating relatively poor performance. After incorporating the dynamic feature $P_{EDG}$, the Dice score improved to 90.87\%, HD95 decreased to 4.42, and TCD improved to 0.0078.This demonstrates that the addition of dynamic learning helps mitigate the effects of motion artifacts and speckle noise, thereby enhancing segmentation accuracy. Compared to the baseline method, the complete DyL-UNet model achieved a 3.91\% improvement in Dice coefficient and a 35.4\% improvement in TCD. This significant enhancement validates the effectiveness of the synergistic interaction between phase constraints and dynamic features. The scatter plot in Figure~\ref{fig:ablation} more intuitively demonstrates that the complete DyL-UNet framework achieves higher segmentation accuracy and better temporal stability across all cardiac structures (LV, LVM, LA).

\section{Conclusion}
This paper proposes a DyL-Unet framework based on dynamic learning for temporal consistency in the segmentation of echocardiography video data, effectively mitigating the issue of inter-frame temporal instability caused by deformation and speckle noise in echocardiography segmentation. By integrating EDG constructed through dynamic learning with cardiac phase information, the CPDA module imposes temporal consistency constraints during segmentation. This enables the network to fully leverage cardiac dynamic features, enhancing segmentation temporal stability. Experiments based on CAMUS and Echo-Dynamic demonstrate that this method achieves outstanding segmentation performance, particularly exhibiting significant superiority over existing methods in temporal consistency. It provides a solution for automated clinical echocardiography analysis that balances accuracy and stability. Future work will focus on enhancing generalization and clinical translation for three-dimensional and multi-view ultrasound applications.
\begin{table}[!t]
\caption{Ablation studies of diffrent modules and methods in DyL-Unet framework.}\label{table:ablation}
\centering
\begin{tabular}{cccc}
\Xhline{1.2pt}
Method & Dice($\uparrow$) & HD95($\downarrow$) & \makecell{Average \\TCD($\downarrow$)}\\
\hline
\makecell{Baseline\\(w/o phase + EDG)} & 88.26 & 5.84 & 0.0096\\
\makecell{DyL-Unet\\(w/o phase)} & 90.87 & 4.42 & 0.0078\\
DyL-Unet & 92.17 & 3.90 & 0.0062\\

\Xhline{1.2pt}
\end{tabular}
\end{table}

\newpage

\bibliographystyle{IEEEbib}
\bibliography{main}

\end{document}